\documentclass[conference]{IEEEtran}
\IEEEoverridecommandlockouts
\usepackage{cite}
\usepackage{amsmath,amssymb,amsfonts}
\usepackage{algorithm}
\usepackage{algorithmic}
\usepackage{graphicx}
\usepackage{textcomp}
\usepackage{xcolor}
\usepackage{url}
\def\BibTeX{{\rm B\kern-.05em{\sc i\kern-.025em b}\kern-.08em
    T\kern-.1667em\lower.7ex\hbox{E}\kern-.125emX}}
\begin{document}

\makeatletter
\newcommand\fs@norules{\def\@fs@cfont{\bfseries}\let\@fs@capt\floatc@ruled
  \def\@fs@pre{}%
  \def\@fs@post{}%
  \def\@fs@mid{\kern3pt}%
  \let\@fs@iftopcapt\iftrue}
\makeatother
\floatstyle{norules}
\restylefloat{algorithm}

\title{Language Detection for Transliterated Content}

\author{\IEEEauthorblockN{Dr. Selva Kumar S }
\IEEEauthorblockA{\textit{Department of Computer Science and Engineering} \\
\textit{B. M. S. College of Engineering}\\
Bangalore, India \\
selva.cse@bmsce.ac.in}
\and
\IEEEauthorblockN{Afifah Khan Mohammed Ajmal Khan}
\IEEEauthorblockA{\textit{Department of Computer Science and Engineering} \\
\textit{B. M. S. College of Engineering}\\
Bangalore, India \\
afifah.cs20@bmsce.ac.in}
\and
\IEEEauthorblockN{Chirag Manjeshwar}
\IEEEauthorblockA{\textit{Department of Computer Science and Engineering} \\
\textit{B. M. S. College of Engineering}\\
Bangalore, India \\
chirag.cs20@bmsce.ac.in}
\and
\IEEEauthorblockN{Imadh Ajaz Banday}
\IEEEauthorblockA{\textit{Department of Computer Science and Engineering} \\
\textit{B. M. S. College of Engineering}\\
Bangalore, India \\
imadh.cs20@bmsce.ac.in}
}
\maketitle

\begin{abstract}

In the contemporary digital era, the Internet functions as an unparalleled catalyst, dismantling geographical and linguistic barriers, particularly evident in texting. This evolution facilitates global communication, transcending physical distances and fostering dynamic cultural exchange. A notable trend is the widespread use of transliteration, where the English alphabet is employed to convey messages in native languages, posing a unique challenge for language technology in accurately detecting the source language. This paper addresses this challenge through a dataset of phone text messages in Hindi, and Russian transliterated into English, utilizing BERT for language classification and Google Translate API for transliteration conversion. The research pioneers innovative approaches to identify and convert transliterated text, navigating challenges in the diverse linguistic landscape of digital communication. Emphasizing the pivotal role of comprehensive datasets for training Large Language Models (LLMs) like BERT, our model showcases exceptional proficiency in accurately identifying and classifying languages from transliterated text. With a validation accuracy of 99\%, our model's robust performance underscores its reliability. The comprehensive exploration of transliteration dynamics, supported by innovative approaches and cutting-edge technologies like BERT, positions our research at the forefront of addressing unique challenges in the linguistic landscape of digital communication. Beyond contributing to language identification and transliteration capabilities, this work holds promise for applications in content moderation, analytics, and fostering a globally connected community engaged in meaningful dialogue.
\end{abstract}

\begin{IEEEkeywords}
Transliteration, Language Identification, Digital Communication, Large Language Models (LLMs), BERT Architecture, Internet Communication.

\end{IEEEkeywords}

\section{Introduction}

In the contemporary digital landscape, the Internet acts as an unparalleled facilitator, dismantling both geographical and linguistic barriers. This transformation is particularly evident in the realm of texting, where individuals from varied language backgrounds converge to communicate and effortlessly exchange information. Through the simplicity of text-based communication, the Internet has transcended physical distances, enabling a seamless flow of ideas and conversations. This evolution in texting has not only connected people globally but has also catalyzed a dynamic cultural exchange. The convergence of diverse linguistic expressions in digital conversations highlights the Internet's role as a universal platform, fostering a global community engaged in continuous dialogue and information sharing.

One striking phenomenon that has gained immense popularity is the practice of transliteration, where users employ the English alphabet to type in their native languages. For example, a Hindi speaker might type "how are you" as "ap kaise ho," using English characters to represent Hindi words. This practice has become ubiquitous, offering a convenient means for multilingual communication. However, this phenomenon poses a unique challenge in the realm of language technology - the need to accurately detect the source language of such transliterated text. The importance of this challenge lies in the potential to enhance user experience and accessibility.

Our research focuses on a dataset\cite{b1} comprising phone text messages, emails, and online forum posts. The selected languages for investigation include Hindi and Russian transliterated into English. To achieve language classification post-transliteration, we employed BERT, a powerful language model. BERT was utilized to classify input text in English alphabet into the correct language label, distinguishing between Chinese, Hindi, and Russian. For translation purposes, the Google Translate API was employed to convert transliterated Russian and Hindi into standard Russian and Hindi characters, ensuring a comprehensive exploration of language dynamics in the digital landscape. This research paper aims to contribute innovative approaches for identifying and converting transliterated text, addressing the unique challenges posed by the diverse linguistic landscape of digital communication.

\begin{figure*}[htbp]
\centerline{\includegraphics[width=16cm]{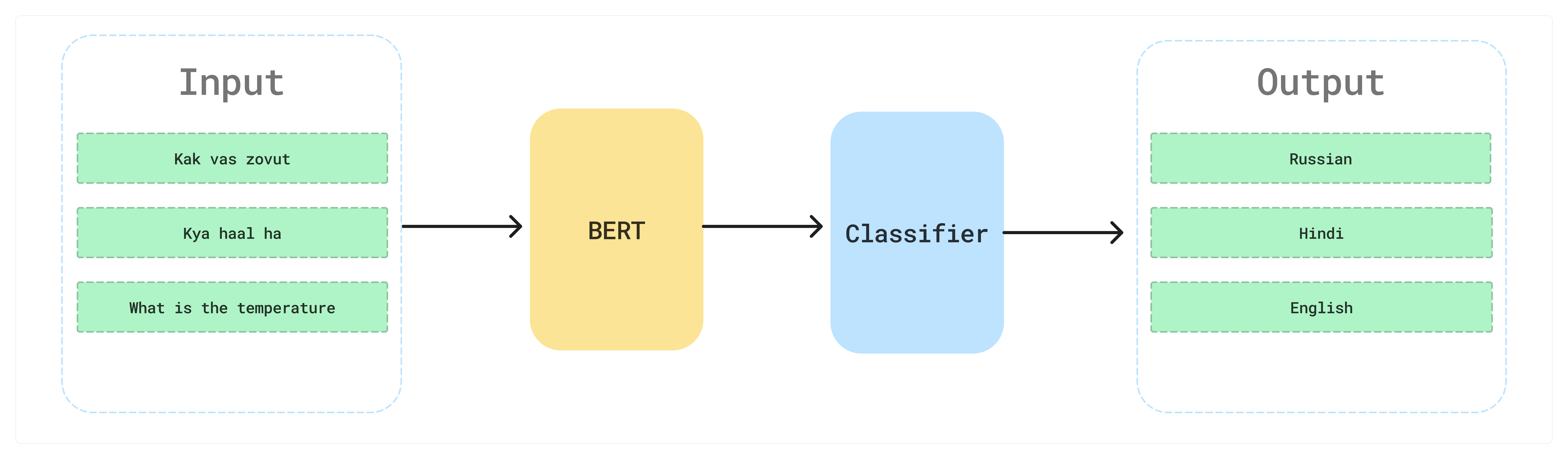}}
\caption{Model Architecture}
\label{fig1}
\end{figure*}

\section{Related Work}
Transliteration-based language models have recently gained significant attention within the realm of artificial intelligence. Noteworthy research papers contribute to our understanding of this domain :

* K Shwait et al. in their paper \cite{b2} propose a comprehensive system for text detection and recognition in multilingual signboards, utilizing PSENet and PyTesseract. Integrating a seq2seq encoder-decoder model, the system successfully transliterates the detected text instances into English, showcasing robustness in handling various shapes and closely packed instances in natural images. The approach prioritizes simplicity in model complexity, demonstrating effectiveness, especially in cases of limited dataset and less intricate transliteration scenarios. The next paper \cite{b3} explores the domain of transliteration models which introduces a highly effective multilingual transliteration model based on modified transformers, showcasing superior performance, particularly for Indic languages transliterated into other Indic languages. Achieving a top-1 accuracy score of 80.7\%, surpassing state-of-the-art models by 29.5\%, the model emphasizes phonetic accuracy, demonstrating its applicability in overcoming language barriers for communication tasks, aligning with our goal of enhancing transliteration and language identification in diverse digital contexts. Another study \cite{b4} by A Gahoi et al. addresses code-mixed machine translation, focusing on English + Hindi to Hinglish and Hinglish to English, achieving top ROUGE-L and WER scores for the first task using mBART with transliteration. The second task, translating Hinglish to English, involves fine-tuning Salesken.AI's pre-trained model, surpassing baseline but with room for further improvement. This next paper\cite{b5} proposed a method that addresses the challenge of transliterating proper names and technical terms from English to Kannada/Telugu for Cross-Language Information Retrieval (CLIR). It introduces a machine learning-based approach, utilizing bilingual proper name lists to automate the transliteration process, showcasing high precision and recall rates. R Nanayakkara et al. propose in their paper \cite{b6} a back-transliteration system for Romanized-Sinhala texts using an NMTR model, achieving favorable results in the context of WhatsApp messages and reasonable performance in Wikipedia texts. The proposed BLSTM-LSTM encoder-decoder model demonstrates effectiveness in addressing the challenges of context-sensitive transliteration rules. Another interesting study \cite{b7} address the challenge faced by non-native individuals in reading Kannada text in Karnataka, proposing a mobile app that utilizes Tesseract SDK for Optical Character Recognition (OCR) and the LIBINDIC Soundex Algorithm for Kannada to English transliteration. The approach aims to enhance communication between diverse cultures, offering a user-friendly and cost-effective solution for real-world image processing, with potential future improvements in error reduction and processing time. A study proposed by in \cite{b8} investigates language identification challenges in social media, focusing on short and noisy tweets. It employs a general-purpose language identification model to semi-automatically label a large tweet corpus, emphasizing the impact of transliterated languages like Arabic and Russian, ultimately demonstrating improved accuracy through an adapted training corpus containing micro-blogging messages. This next study proposed by * MU Athukorala et al. \cite{b9} introduces "Swa Bhasha" a novel system for Singlish to Sinhala transliteration using a rule-based approach and a numeric coding system. With an 84\% accuracy at the word level and 92\% accuracy in suggestion-level predictions, it enhances the Sinhala users' experience, addressing challenges in modern texting with Singlish words. Lastly, a paper proposed by * PJ Antony et al. \cite{b10} on English to Kannada transliteration using structured output Support Vector Machines (SVM) and sequence labeling. The proposed approach achieves exact Kannada transliterations for 87.28\% of English names, demonstrating its potential for applications like machine translation and cross-language information retrieval.
\section{Need for the Model} 

Certain online content is in languages other than English but is written using the English alphabet. For instance, on country-specific forums like the subreddit for India, individuals often compose posts in their native language, such as Hindi, but use the English alphabet due to typing on an English keyboard. To effectively handle this type of data, it is crucial to identify the language, transliterate it into the original script, and then translate it. Our model specifically focuses on the language detection aspect of this process. Through the use of our model, a broader range of internet data becomes available for streamlined data processing.

\section{Implementation} \label{modelSection}

\subsection{Training Dataset Generation} 
\label{generation}
\begin{figure}
    \centering
    \includegraphics[width=0.5\linewidth]{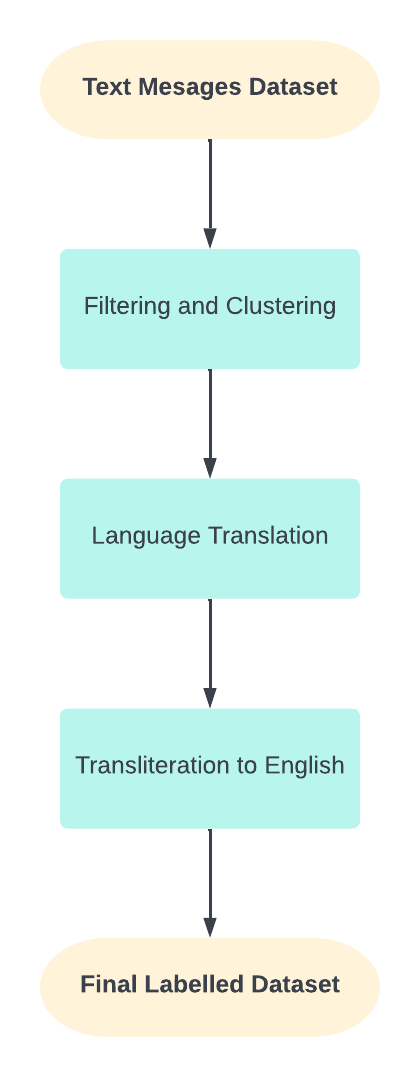}
    \caption{Dataset Generation}
    \label{fig_dataset}
\end{figure}
The training dataset was created following the procedure outlined in Fig. \ref{fig_dataset}. Text messages were selected from \cite{b11}, which encompasses various categories corresponding to distinct devices from which the messages originated. These categories were consolidated into a single dataset and subjected to filtration. Following filtration, the text messages underwent translation into Russian and Hindi using the Google Translate API \cite{google_translate_api}. Subsequently, the messages were transliterated into the English alphabet using the Polyglot Library \cite{polyglot_library}.

\subsection{Model} 
\begin{figure}[htbp]
\centerline{\includegraphics[width=5cm]{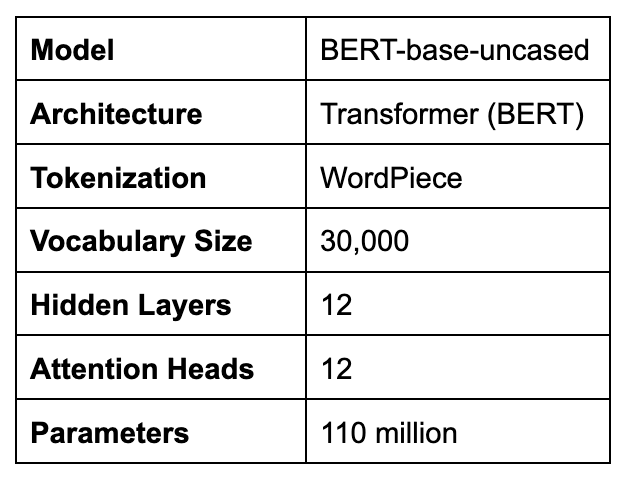}}
\caption{Model Parameters}
\label{model_table}
\end{figure}
Our model as illustrated in Fig \ref{fig1} is designed to seamlessly process input in any language written in the English alphabet, employing the state-of-the-art BERT architecture. Through meticulous training from scratch, our BERT model has been fine-tuned to accurately identify and classify diverse languages. This training process involved exposing the model to a wide range of multilingual text datasets, enabling it to learn intricate language patterns and distinctions. As a result, our model excels at the task of language identification, dynamically recognizing the language of the input text and categorizing it accordingly. This robust capability allows our model to cater to a broad spectrum of linguistic diversity, making it a versatile and effective solution for language-based applications. The model parameters are described in Fig \ref{model_table}.

\section{Experimental Setup And Evaluation}

Our experimental setup and training process are detailed below:

\subsection{Dataset}
We utilized 3,000 sentence-language pairs generated through the procedure outlined in Fig \ref{generation}, partitioning the dataset into 80\% for training and 20\% for validation.

\subsection{Training}
The model was trained using the Adam W optimizer with a learning rate set to $5 \times 10^{-5}$ for 5 epochs. The training utilized a batch size of 4, and the process was performed on a T4 GPU.

\subsection{Evaluation}
\begin{figure}[htbp]
\centerline{\includegraphics[width=7cm]{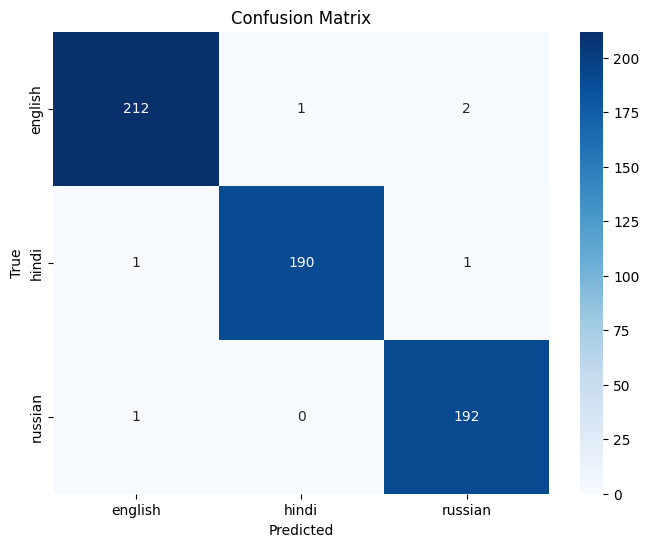}}
\caption{Confusion Matrix}
\label{confusion}
\end{figure}
\begin{figure}[htbp]
\centerline{\includegraphics[width=5cm]{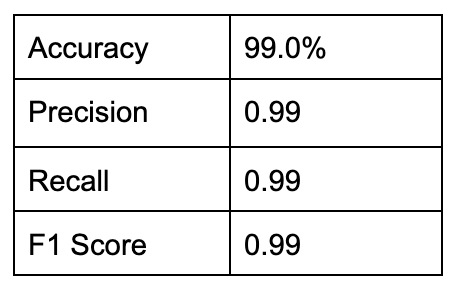}}
\caption{Evaluation Metrics}
\label{accuracy}
\end{figure}

The validation dataset achieved an accuracy of 99\%.
\begin{figure}
    \centering
    \includegraphics[width=0.9\linewidth]{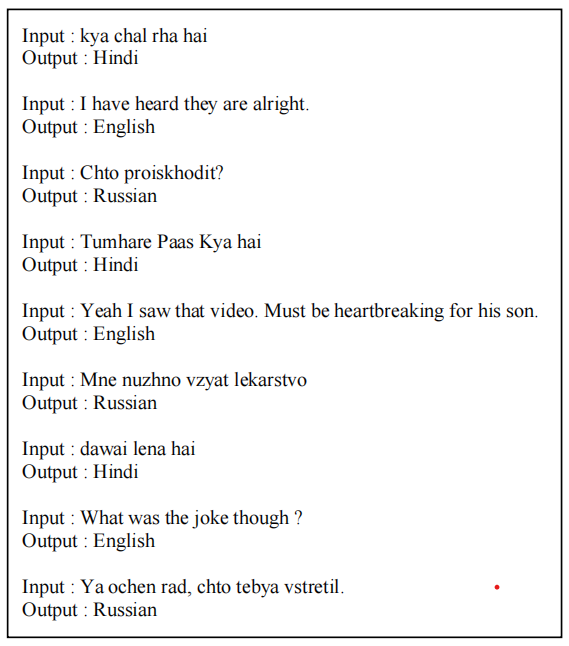}
    \caption{Sample Inference}
    \label{inference}
\end{figure}
Fig. \ref{inference} shows some sample inferences. Fig. \ref{confusion} shows the confusion matrix obtained for the validation dataset, and Fig. \ref{accuracy} shows the validation metrics.

\section{Conclusion}

In conclusion, our research explores the transformative impact of transliteration-based language models in the contemporary digital landscape. The Internet, acting as a global facilitator, has seen the convergence of diverse linguistic backgrounds through text-based communication. Transliteration, where users employ the English alphabet to represent their native languages, has become a prevalent practice fostering multilingual communication. Our model, built on the advanced BERT architecture, demonstrates exceptional proficiency in accurately identifying and classifying languages from text transliterated into the English alphabet. Through experimentation and evaluation, we achieved a validation accuracy of 99\%, showcasing the model's robust performance. The comprehensive exploration of transliteration dynamics, supported by our innovative approaches and the incorporation of cutting-edge technologies like BERT, positions our research at the forefront of addressing the unique challenges posed by the diverse linguistic landscape of digital communication. This work not only contributes to language identification and transliteration capabilities but also holds promise for applications in content moderation, analytics, and fostering a globally connected community engaged in meaningful dialogue.

\end{document}